\documentclass[final]{cvpr}

\usepackage{times}
\usepackage{epsfig}
\usepackage{graphicx}
\usepackage{amsmath}
\usepackage{amssymb}
\usepackage{multirow}
\usepackage{multicol}
\usepackage{booktabs}
\usepackage{caption}
\usepackage{mathtools}
\usepackage{enumerate}
\usepackage{eufrak}

\usepackage[pagebackref=true,breaklinks=true,colorlinks,bookmarks=false]{hyperref}

\begin{document}

\title{
Seeing Out of tHe bOx: \\End-to-End Pre-training for Vision-Language Representation Learning
}

\author{
Zhicheng Huang$^{1,2}$\thanks{Equal Contribution. This work was performed when Zhicheng Huang, Zhaoyang Zeng and Yupan Huang were visiting Microsoft Research Asia as research interns.},
Zhaoyang Zeng$^{3}$\footnotemark[1],
Yupan Huang$^{3}$\footnotemark[1],
Bei Liu$^{4}$,
Dongmei Fu$^{1,2}$,
Jianlong Fu$^{4}$\\
$^1$School of Automation and Electrical Engineering, University of Science and Technology Beijing\\ $^2$Beijing Engineering Research Center of Industrial Spectrum Imaging\\ $^3$Sun Yat-sen University, $^4$Microsoft Research Asia
}

\maketitle

\begin{abstract}
\vspace{-2mm}
    We study joint learning of Convolutional Neural Network (CNN) and Transformer for vision-language pre-training (VLPT) which aims to learn cross-modal alignments from millions of image-text pairs. State-of-the-art approaches extract salient image regions and align regions with words step-by-step. As region-based visual features usually represent parts of an image, it is challenging for existing vision-language models to fully understand the semantics from paired natural languages. In this paper, we propose \textbf{SOHO} to ``\textbf{S}ee \textbf{O}ut of t\textbf{H}e b\textbf{O}x'' that takes a whole image as input, and learns vision-language representation in an end-to-end manner. SOHO does not require bounding box annotations which enables inference 10 times faster than region-based approaches. In particular, SOHO learns to extract comprehensive yet compact image features through a visual dictionary (VD) that facilitates cross-modal understanding. VD is designed to represent consistent visual abstractions of similar semantics. It is updated on-the-fly and utilized in our proposed pre-training task Masked Visual Modeling (MVM). We conduct experiments on four well-established vision-language tasks by following standard VLPT settings. In particular, SOHO achieves absolute gains of 2.0\% R@1 score on MSCOCO text retrieval 5k test split, 1.5\% accuracy on NLVR$^2$ test-P split, 6.7\% accuracy on SNLI-VE test split, respectively. 
\end{abstract}

\vspace{-6mm}
\section{Introduction}
\vspace{-1mm}

\begin{figure}
  \includegraphics[width=0.95\linewidth]{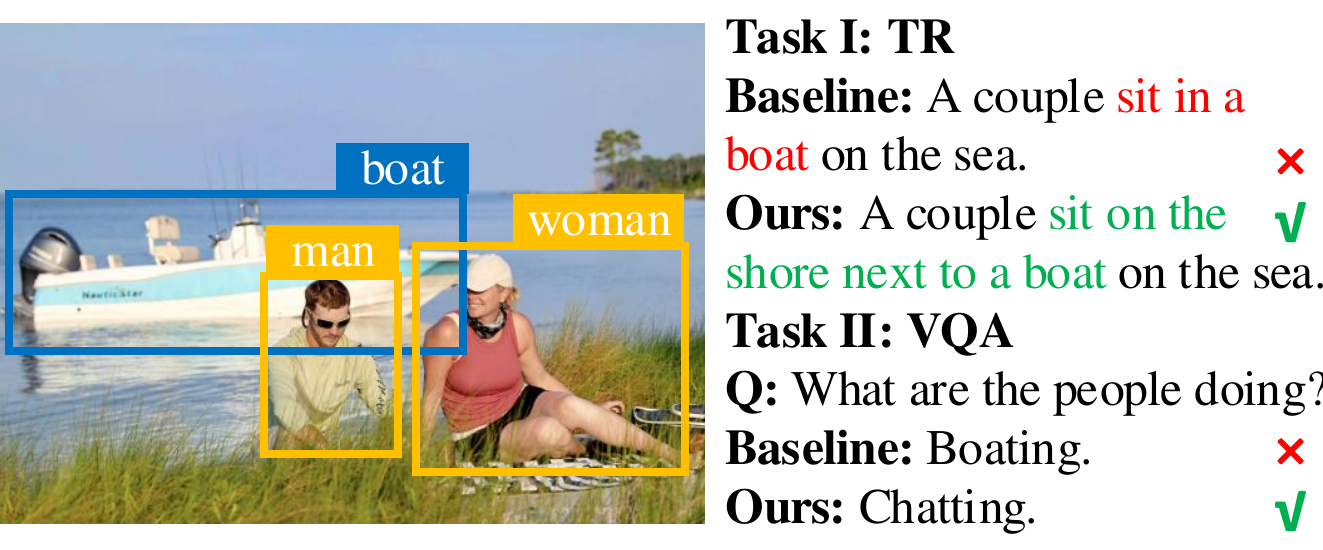}
  \vspace{-4mm}
  \caption{Comparisons of SOHO and region-based methods by top-1 image-to-text retrieval (TR) and visual question answering (VQA) results. Baselines lack global context and fail to understand the image. SOHO discovers visual clues out of region boxes and infers correct human activities. [Best viewed in color.]}
  \label{fig:ob_pt_ct}
  \vspace{-7mm}
\end{figure}

With the success of Transformer and self-supervised learning, we have recently witnessed a boosting number of research works on cross-modal learning, especially on vision-language pre-training (VLPT) \cite{chen2020uniter,li2019unicoder,li2019visualbert,lu2019vilbert,su2019vl,tan2019lxmert,zhou2019unified}. 
VLPT models learn better cross-modal representation with large-scale easy-accessible image-text pairs. They have established state-of-the-art results in many vision-language tasks, such as visual question answering (VQA) \cite{Antol_2015_ICCV}, image-text retrieval \cite{lin2014microsoft}, natural language for visual reasoning (NLVR) \cite{suhr2019corpus}, etc. 
Visual representation plays an important role in VLPT models.
The recent success of VLPT models has been accompanied by the usage of region-based image features, which are extracted by object detectors pre-trained on the Visual Genome dataset~\cite{Anderson_2018_CVPR}.
However, there are three challenges to directly utilize region-based image features for vision-language understanding.
Firstly, regions focus on objects inside bounding boxes while neglecting the contextual information out of the boxes, which is important for relation understanding and reasoning. For example in Figure \ref{fig:ob_pt_ct}, we can easily detect ``man'', ``woman'' and ``boat'' in the image. However, without the contextual information out of these boxes, a model will misunderstand the relation as ``people boating'' and result in an incorrect answer for either text retrieval or VQA. 
Secondly, visual understanding of images will be limited to the pre-defined categories for regions.
Thirdly, most region-based image features are extracted by a detection model, which will suffer from low quality, noise, and over-sampling \cite{Anderson_2018_CVPR} and rely on large-scale boxes annotation data. Although some works try to train detection model\cite{2017Multiple,zeng2019wsod2} with weakly-supervised, the performance is far below the requirements. 
Recently, some works challenge that
grid-based convolutional features
are also effective to learn visual representations~\cite{desai2021virtex,huang2020pixel,jiang2020defense,sariyildiz2020icmlm}.
Among them, Jiang \textit{et al.} show that grid features can be equally effective as region features for VQA~\cite{jiang2020defense}.
Sariyildiz \textit{et al.} and Desai \textit{et al.} use image-text data to train visual backbone for recognition tasks (e.g., image classification) ~\cite{desai2021virtex,sariyildiz2020icmlm}.
These models are designed either for specific vision-language task~\cite{jiang2020defense} or vision task~\cite{desai2021virtex,sariyildiz2020icmlm}.
In this paper, we focus on VLPT and propose one of the first end-to-end VLPT model without relying on region features.

To overcome the limitation of region-based image features and
better utilize image-text pairs for cross-modal understanding, we propose \textbf{SOHO}, an end-to-end vision-language pre-training framework to directly learn image embedding, language embedding, and their semantic alignment from image-text pairs.
Compared with existing VLPT works, SOHO adopts a simple pipeline that does not require a complicated visual backbone for pre-training and releases the design effort for VLPT tasks. Without the requirement of laborious annotated categories or boxes, SOHO can enrich visual semantics by directly optimizing visual representations by a wider range of image-text data.

End-to-end learning for vision and language raises challenges by different representations of the two modalities. Visual representation at pixel-level is much more diverse and dense than language embedding. And the lack of explicit supervision for pixel-level language adds the difficulty to alignment learning. To tackle the above problems, we introduce a visual dictionary (VD) which represents more comprehensive and compact semantics in visual domain. 
To learn the visual dictionary, we design a moving-averaged encoder to group visual pixels with similar visual semantics. VD can be dynamically updated through our trainable CNN backbone directly from visual-language data during pre-training. For pre-training tasks, we propose a novel Masked Vision Modeling (MVM) based on the learned visual dictionary besides two commonly used tasks, Masked Language Modeling (MLM) and Image-Text Matching (ITM).

Our contributions are summarized as follows:
\textit{(\romannumeral1)} We propose SOHO, one of the first end-to-end VLPT models to learn cross-modal representation directly with image-text pairs. Without the need of extracting bounding boxes, our model can achieve at least 10 times speedup for inference.
\textit{(\romannumeral2)} To better align visual features and language tokens, we propose a novel dynamic-updated visual dictionary that represents a visual abstraction of similar semantics in images. 
\textit{(\romannumeral3)} We conduct extensive experiments on four well-established downstream tasks. 
Experimental results show that SOHO can improve the SOTA performance with absolute gains of 2.0\% R@1 score on MSCOCO text retrieval 5k test split, 1.5\% accuracy on NLVR$^2$ test-P split, 6.7\% accuracy on SNLI-VE test split, and 0.56\% VQA score on VQA2.0 test-std split.
We will release both model and code to facilitate the research community\footnote{https://github.com/researchmm/soho}.

\vspace{-2mm}
\section{Related Work}\label{section:related_work}
{\subsection{Visual Representation for Vision-Language}
Visual representation learning for vision-language understanding is a long-standing research topic.
Early works use CNN classification models pre-trained on ImageNet to extract visual features~\cite{deng2009imagenet,li2018tell,liu2018beyond,NIPS2016_9dcb88e0,yang2016stacked,yu2017multi}.
Later on, Anderson \textit{et al.} propose a Bottom-Up and Top-Down Attention (BUTD) detection model~\cite{Anderson_2018_CVPR} pre-trained on Visual Genome dataset to extract salient region features as visual inputs for VQA and image captioning tasks.
The BUTD features are adopted by many vision-language works~\cite{Anderson_2018_CVPR,krishna2017visual,singh2018pythia,suhr2019corpus} and pre-training works~\cite{chen2020uniter,kim2018bilinear,tan2019lxmert}.
Recently, some works propose to directly learn visual representations in the form of grid features with convolutional networks in specific vision-language tasks~\cite{jiang2020defense} or vision recognition tasks~\cite{desai2021virtex,sariyildiz2020icmlm}.
Our work shares a similar format of visual representation with~\cite{jiang2020defense} while we focus on the area of vision-language pre-training and propose {the first end-to-end VLPT model} without relying on the box annotations.
}

{
VideoBERT~\cite{sun2019videobert} and the bag of words~\cite{fei2005bayesian} literature also use vector quantization to represent visual information.
The key difference between VD and related works is that we dynamically update the VD-based embedding with the output of a trainable visual encoder, instead of pre-computed input features.
The dynamic updating mechanism for VD can capture text-guided semantics from the vision-language dataset.
Thus the model can be directly optimized with high-level semantics for VL understanding and alignment.
}

\begin{figure*}
\begin{center}
\includegraphics[width=0.95\linewidth]{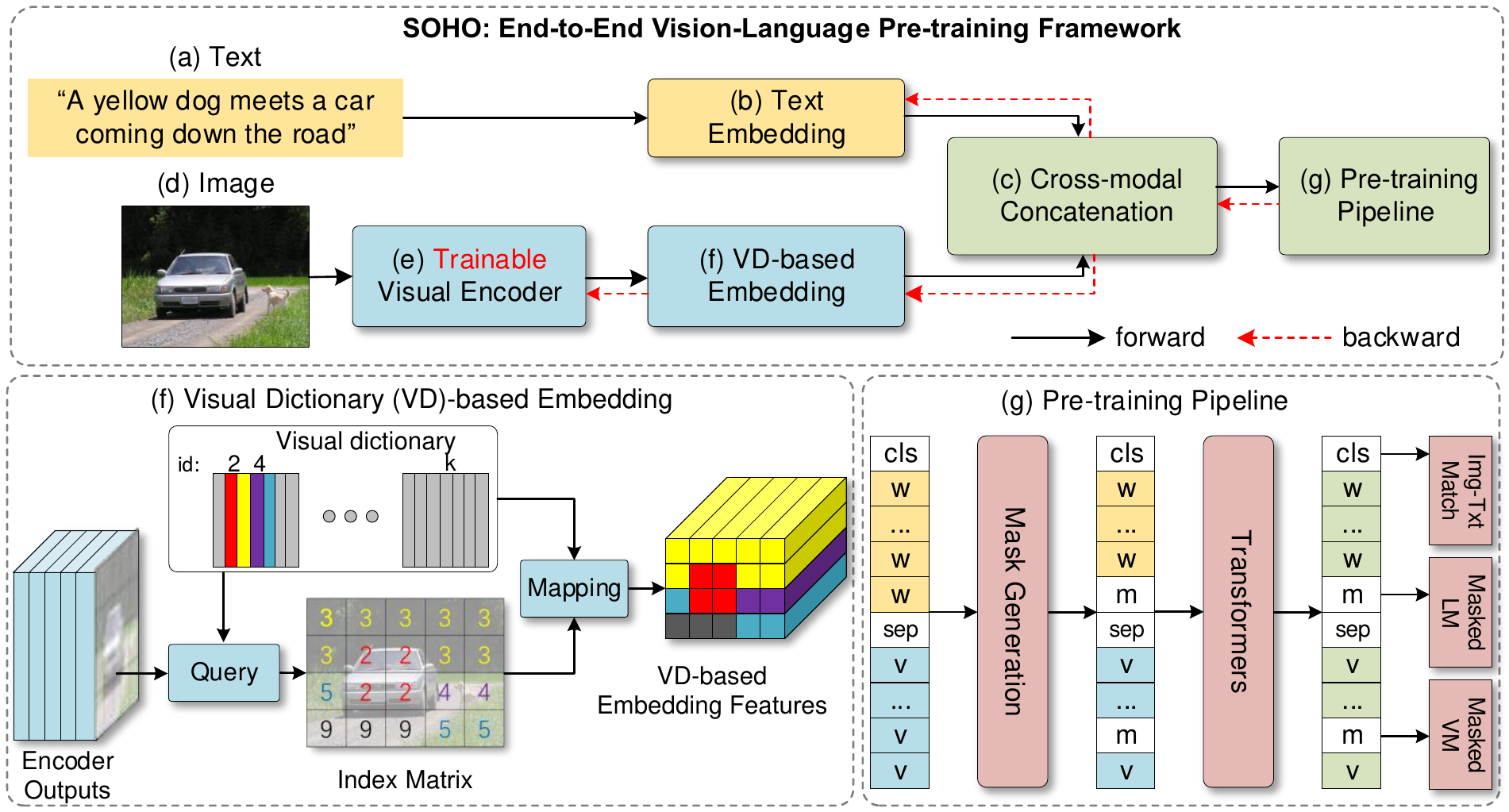}
\end{center}
\vspace{-6mm}
   \caption{The framework of the proposed end-to-end pre-training model SOHO. For an input text (a), we use the text embedding operation (b) to extract the textual embedding features. For an input image (d), we propose to use a trainable CNN-based encoder (e) to extract visual representations. To further transform image features to consistent semantics, we apply a visual dictionary-based image embedding (f) to the image encoder outputs. Finally, we apply multi-layer Transformers to the output of multi-modal concatenation (c) with three pre-training tasks. Note that the index matrix in (f) will be used as labels in the masked VM task in (g). [Best viewed in color.]}
\label{fig:framework}
\vspace{-6mm}
\end{figure*}

\subsection{Pre-training for Vision-Language}
\vspace{-1mm}

{
Many vision-language pre-training (VLPT) works have been proposed to learn cross-modal representations~\cite{chen2020uniter,li2019unicoder,li2019visualbert,lu2019vilbert,su2019vl,sun2019videobert,tan2019lxmert,zhou2019unified}.
They can be categorized as two-stream or single-stream models.
The two-stream models process visual and language information respectively and fuse them afterward by another Transformer layer~\cite{lu2019vilbert,tan2019lxmert}.
} 
Contrarily, the single-stream models use BERT~\cite{devlin2018bert} to learn a bi-directional joint distribution over the detection bounding box feature and text embedding feature~\cite{alberti2019fusion,chen2020uniter,li2019unicoder,li2019visualbert,su2019vl,zhou2019unified}.
Both types use the Transformer-based model to learn vision-language joint embedding features. While they neglect that visual representation learning is also important to vision-language tasks.

The key differences between our SOHO and existing VLPT works are
1) SOHO adopts a simple VLPT pipeline.
Our vision backbone only uses ImageNet pre-trained parameters, and achieves even higher performance than existing VLPT works using VG features on five downstream tasks. 
2) SOHO uses the least annotations to achieve SOTA performances.
3) SOHO enriches visual semantics by directly optimizing visual inputs for target language tasks.

\vspace{-2mm}
\section{Approach}
The overall architecture of our proposed vision-language pre-training framework SOHO is shown in Figure~\ref{fig:framework}. SOHO is an end-to-end framework, which consists of a trainable CNN-based visual encoder, a visual dictionary (VD) embedding module, and a multi-layer Transformer. The visual encoder takes an image as input and produces the visual features. 
VD embedding module is designed to aggregate diverse visual semantic information into visual tokens with a proposed visual dictionary.
The Transformer is adopted to fuse features from visual and language modalities, and produce task-specific output. SOHO can be end-to-end pre-trained by Masked Vision Modeling (MVM), Masked Language Modeling (MLM), and Image-Text Matching (ITM) tasks. SOHO can also be easily adapted to several downstream tasks including Image-Text Retrieval, VQA, NLVR, and Visual Entailment.

\vspace{-2mm}
\subsection{Trainable Visual Encoder}
\label{sec:3.1}
\vspace{-1mm}
Most recent vision-language researches follow Bottom-up and Top-Down attention~\cite{Anderson_2018_CVPR} to extract region-level visual features by a Faster R-CNN~\cite{ren2015faster} detector which is pre-trained on the Visual Genome~\cite{krishna2017visual} dataset. 
The representation ability of such extracted region-based features will be limited by the pre-defined object and attribute categories (i.e. $1,600$ objects and $400$ attributes). Besides, some contextual information out of the regions is important for VL understanding, while being neglected because they are out of the pre-defined categories. Even though considering the whole image as a region and extract its feature as global representation is an improved solution, the detector can not guarantee the feature quality because such global regions are unseen in the training stage. Despite that, most recent VLPT models adopt pre-extracted region-level visual features because it is complicated to end-to-end fine-tune an object detector in VL tasks. Besides, the extracted region-level visual features have a semantic gap with the language domain, while existing works try to bridge such domain gap by only one or several fully-connected layers.

To keep all visual information, we propose to use a trainable CNN visual encoder, which takes the whole image as input and produces image-level visual features instead of region-level features. Without the limitation of bounding boxes, the visual encoder can be end-to-end learned and updated from pre-training losses or downstream task-specific losses, and further optimize the cross-modal learning in turn.
Given an input image $\mathcal{I}$, we get its feature $\mathcal{V}$ by:
\vspace{-2mm}
\begin{equation}
\mathcal{V} = E(\mathcal{I}, \theta) \in \mathbf{R}^{l\times{c}},
\vspace{-2mm}
\end{equation}
where $E(\cdot, \theta)$ is the visual feature encoder with parameter $\theta$. $l$ denotes the number of embedded feature vectors, and $c$ is the embedded dimension. We adopt ResNet~\cite{he2016deep} pre-trained on ImageNet~\cite{deng2009imagenet} followed by a $1\times 1$ convolutional layer and a $2\times 2$ max pooling layer as the architecture of the encoder $E$. For simplicity, we use $v_i$ to denote the $i^{th}$ feature vector of $\mathcal{V}$ for the rest of this paper.

\vspace{-1mm}
\subsection{Visual Dictionary}
\label{sec:3.2}\label{subsection:visual_dictionary}
\vspace{-1mm}

The visual feature $\mathcal{V}$ extracted by visual feature encoder is more diverse and dense than language word tokens, which will bring difficulty to the learning of cross-modal understanding. To bridge its representation gap from language tokens, we propose a visual dictionary (VD) to tokenize the visual features by aggregating similar visual semantic into the same image feature.

\noindent\textbf{Visual Dictionary Embedding.} 
We define a visual dictionary (VD) as a matrix $\mathcal{D}\in\mathbf{R}^{k\times{c}}$ which contains $k$ embedding vectors with $c$-dim. The $j^{th}$ embedding vector is denoted as $d_j$. For visual feature $v_i$, we compute it mapping index by searching nearest neighbor in $\mathcal{D}$, denoted as:
\vspace{-3mm}
\begin{equation}
    h_i =\text{argmin}_{j}\lVert v_{i}-d_{j}\rVert_{2}.
\vspace{-2mm}
\end{equation}
We define visual dictionary embedding as a mapping function $f$, which maps $v_i$ to $\mathcal{D}$ by:
\vspace{-2mm}
\begin{equation}
    f(v_i) = d_{h_i},
    \label{eq:3}
\vspace{-2mm}
\end{equation}
which uses the nearest embedding vector to represent the visual feature. We denote $f^{-1}(j)$ as an inverse mapping function, which maps the index $j$ back to a group of visual features. We use $|f^{-1}(j)|$ to represent the inverse mapping group size, and use $f(\mathcal{V})$ to represent the encoding features.

\noindent \textbf{Momentum Learning for Visual Dictionary Update.}
The visual dictionary $\mathcal{D}$ is randomly initialized, and further updated by a moving average operation in one mini-batch, which is denoted as:
\vspace{-3mm}
\begin{equation}
\begin{aligned}
    \hat{d_j} =\gamma * d_j + (1-\gamma) * \frac{\sum_{h_i=j}{v_i}}{|f^{-1}(j)|}, \\
    \label{eq:4}
\end{aligned}
\vspace{-8mm}
\end{equation}
where $\hat{d_j}$ indicates the updated embedding vector of $d_j$, and $\gamma$ is a momentum coefficient whose value range is $[0, 1]$. 
Note that Eqn.~\ref{eq:4} will only be applied when $|f^{-1}(j)|\neq 0$.

\noindent \textbf{Gradient Back Propagation.}
Since the $\text{argmin}$ operation is not differentiable, the gradient back propagation will be stopped by the visual dictionary. To make the visual feature encoder trainable, we follow \cite{van2017neural} to update $f(v_i)$ by:
\vspace{-1mm}
\begin{equation}
f(v_i) = sg[d_{h_i} - v_i] + v_i,
\vspace{-1mm}
\end{equation}
where $sg[\cdot]$ is the stop gradient operator.

The visual dictionary performs an online clustering on visual feature maps based on feature similarity, and represents each feature vector by its cluster center. Feature vectors sharing similar semantics will be aggregated into the same cluster, and the clustered index can be considered as a virtual visual semantic label. Since the clustering can be affected by the vision-language learning tasks, the learned semantics of each embedding vector is more suitable for cross-modal understanding and alignment.

The visual dictionary faces a cold-start problem, where directly copying the gradient from randomly initialized embedding vectors to visual feature maps will lead to incorrect model optimization direction (i.e., mode collapse). Therefore, we freeze the parameters of ResNet in the visual feature encoder in the first $10$ training epochs.

\begin{table*}[t]
\small
    \centering
    \caption{Statistics of different tasks. Notation ``*'' denotes Karpathy split~\cite{karpathy2015deep}. 
    Notation ``-'' denotes not applicable.
    Detailed train/test image and text numbers can be found in the supplementary material.
    }
    \vspace{-2mm}
    \begin{tabular}{c|c|c|c|c}
    \hline
    \textbf{Task} & \textbf{Dataset} & \textbf{Train Split} & \textbf{Test Split} & \textbf{Metric} \\
    \hline
    \multirow{2}{*}{Pre-training} & VG~\cite{krishna2017visual} & train & - & - \\
    & MSOCO~\cite{lin2014microsoft} & train+restval* & - & - \\
    \hline
    \multirow{2}{*}{Image-Text Retrieval} & MSCOCO~\cite{lin2014microsoft} & train+restval* & test* & \multirow{2}{*}{Recall@1,5,10}  \\
    & Flickr30K~\cite{plummer2015flickr30k} & train & test* & \\
    \hline
    Visual Question Answering & VQA2.0~\cite{goyal2017making} & train+val & test-dev/test-std & VQA-score~\cite{goyal2017making} \\
    \hline
    Visual Reasoning & NLVR$^2$~\cite{suhr2019corpus} & train & dev/test-P & Top-1 Accuracy \\
    \hline
    Visual Entailment & SNLI-VE~\cite{xie2018visual} & train & val/test & Top-1 Accuracy \\
    \hline
    \end{tabular}
\label{table:tasks}
\vspace{-5mm}
\end{table*}

\vspace{-1mm}
\subsection{Pre-training Pipeline}
\label{pretrain}
\vspace{-1mm}
We apply a multi-layer Transformer to learn cross-modal representations with the fusion of visual and language features. In order to learn a universal representation for vision and language-related tasks, we apply the self-supervised method to pre-train the model on a large aggregated dataset. We follow the existing works~\cite{chen2020uniter,li2019unicoder,lu2019vilbert,su2019vl,tan2019lxmert,zhou2019unified} to adopt Masked Language Modeling (MLM) and Image-Text Matching (ITM) pre-training tasks. Besides, we propose a novel Masked Visual Modeling (MVM) pre-training task based on the virtual visual semantic labels produced by the visual dictionary.

\noindent \textbf{Cross-Modal Transformer.}
For visual representation, we utilize 2-D position embedding computed by sine function to encode spatial information of visual tokens following other works \cite{carion2020end,dosovitskiy2020image,parmar2018image}.
For the input sentence, we follow BERT~\cite{devlin2018bert} to tokenize it, and then represent the tokens by embedding vectors $\mathcal{W}$. We use $w_i$ to denote the $i^{th}$ embedding vector in $\mathcal{W}$. The word embedding and the VD embeddings share the dimension $c$ on their outputs. We concatenate the VD embeddings and word embedding vectors together to form an input sequence for cross-modal learning. Similar to other VLPT models, we add two special tokens [\textit{CLS}] and [\textit{SEP}] into the input sequence to indicate classification position and the end of a text, respectively. A multi-layer Transformer is adopted to take the joint vision-language input, and outputs the attended features.

\noindent\textbf{Masked Language Modeling.} We follow \cite{chen2020uniter} and adopt Masked Language Modeling (MLM) to encourage the model to build the mapping between language tokens and visual contents.
The goal of MLM is to predict the masked word tokens based on other word tokens~$\mathcal{W}_{\backslash i}$ and all image features~$f(\mathcal{V})$ by minimizing the negative log-likelihood. The learning target can be formulated as:
\vspace{-3mm}
\begin{equation}
    \mathcal{L}_{\textrm{MLM}}=-\mathbb{E}_{(\mathcal{W},f(\mathcal{V}))\sim D}\log p(w_{i}|\mathcal{W}_{\backslash i},f(\mathcal{V})),
    \label{eq:mlm}
    \vspace{-2mm}
\end{equation}
where $D$ indicate hereinafter the whole training dataset. We adopt the same masking strategy used in BERT~\cite{devlin2018bert}.

\noindent\textbf{Masked Visual Modeling.} We propose Masked Visual Modeling (MVM) by visual dictionary, which is a symmetry to the MLM. We randomly mask the image features before feeding them into the Transformer. The learning target of MVM is denoted as:
\vspace{-2mm}
\begin{equation}
    \mathcal{L}_{\textrm{MVM}}=-\mathbb{E}_{(\mathcal{W},f(\mathcal{V}))\sim D}\log p(f(v_j)|\mathcal{W},f(\mathcal{V})_{\backslash j}).
    \vspace{-2mm}
\end{equation}
The goal of MVM is to predict the masked image features based on their surrounding image features~$f(\mathcal{V})_{\backslash j}$ and all language tokens~$\mathcal{W}$ by minimizing the negative log-likelihood. MVM can encourage the model to infer visual knowledge from the contextual visual information as well as language. When an image feature $v_i$ is masked, its mapping index $h_i$ in VD is considered as its label. In visual feature maps, neighbor features may have similar values, and thus share the same mapping index. This will cause the model to directly copy the label from surrounding features as predictions in a lazy way. To prevent this, in the masking stage, we first randomly select an existing label index $j$, then replace all visual embedding vectors in $f^{-1}(j)$ with the special [\textit{MASK}] token embedding vector.

\noindent\textbf{Image-Text Matching.} To enhance the cross-modal matching, we adopt Image-Text Matching (ITM) task for pre-training as in previous works \cite{chen2020uniter}. 
We apply a binary classifier $\phi(\cdot)$ on the joint embedding feature of [\textit{CLS}] token to predict whether the input image and text are matched or not. ITM task is driven by the following loss function:
\vspace{-2mm}
\begin{equation}
    \mathcal{L}_{\textrm{ITM}}=-\mathbb{E}_{(\mathcal{W},f(\mathcal{V}))\sim D}\log p(y|\phi(\mathcal{W},f(\mathcal{V}))),
    \label{eq:itm}
    \vspace{-2mm}
\end{equation}
where $y\in \{0,1\}$ indicates whether the image and text is matched ($y=1$) or not ($y=0$).

The visual feature encoder, VD-based image embedding module and the cross-modal Transformer is end-to-end jointly trainable. We assign equal loss weight to the three pre-training objectives, and thus the full pre-training objective of SOHO is:
\vspace{-2mm}
\begin{equation}
    \mathcal{L}_{\textrm{Pre-training}}=\mathcal{L}_{\textrm{MLM}}+\mathcal{L}_{\textrm{MVM}}+\mathcal{L}_{\textrm{ITM}}.
    \label{eq:loss}
    \vspace{-1mm}
\end{equation}

\subsection{Pre-training Datasets}
\vspace{-1mm}
Several large-scale datasets have been proposed to facilitate VL pre-training. According to typical settings in UNITER~\cite{chen2020uniter}, these datasets can be categorized into two classes: ``in-domain'' and ``out-domain''. In our work, we use ``in-domain'' as a pre-training dataset as most VL pre-training tasks are built on them~\cite{chen2020uniter,li2019visualbert,tan2019lxmert}. We construct our pre-training datasets with MSCOCO~\cite{lin2014microsoft} and VG~\cite{krishna2017visual}.

To avoid data leak, we only use the \textit{train} and \textit{restval} splits of MSCOCO dataset, and the \textit{train} and \textit{val} splits of VG dataset in the training stage.
The detailed statistic of our pre-training datasets can be found in Table~\ref{table:tasks}.
Detailed comparisons of pre-training dataset usage of most VLPT works, including our train/test image and text numbers, are included in our supplementary material.

\begin{table*}[t]
\small
    \centering
    \caption{Evaluation of image-to-text retrieval (TR) and text-to-image retrieval (IR) task on MSCOCO Dataset. "-" indicates the detail is not reported.}
    \vspace{-3mm}
    \renewcommand\tabcolsep{5pt}
    \begin{tabular}{c|c|cccccc|cccccc}
    	\hline
         \multirow{2}{*}{Model} &\multirow{2}{*}{Backbone} &\multicolumn{3}{c}{TR} & \multicolumn{3}{c|}{IR} &\multicolumn{3}{c}{TR} & \multicolumn{3}{c}{IR}\\
         & & R@1 & R@5 & R@10 & R@1 & R@5 & R@10 & R@1 & R@5 & R@10 & R@1 & R@5 & R@10 \\
         \hline \hline
         \multicolumn{2}{c|}{}&\multicolumn{6}{c|}{1K Test set} &  \multicolumn{6}{c}{5K Test set}\\
         \hline
         VSE++\cite{faghri2017vse++}& R152 & 64.6 & 90.0 & 95.7 & 52.0 & 84.3 & 92.0
         & 41.3 & 71.1 & 81.2 & 30.3 & 59.4 & 72.4\\
         SCAN\cite{lee2018stacked}& R101& 72.7 & 94.8 & 98.4 & 58.8 & 88.4 & 94.8
         & 50.4 & 82.2 & 90.0 & 38.6 & 69.3 & 80.4\\
         \hline
     
         Unicoder-VL\cite{li2019unicoder}& -& 84.3 & 97.3 & {99.3} & 69.7 & 93.5 & 97.2
         & 62.3 & 87.1 & 92.8 & 46.7 & 76.0 & 85.3\\
       
         UNITER\cite{chen2020uniter} & R101 & - & - & - & - & - & -
         & 64.4 & 87.4 & 93.1 & 50.3 & \textbf{78.5} & \textbf{87.2} \\
       
         \hline
       
         SOHO (ours) & R101 & \textbf{85.1} & \textbf{97.4} &\textbf{99.4} & \textbf{73.5} & \textbf{94.5} & \textbf{97.5}
          & \textbf{66.4} & \textbf{88.2} & \textbf{93.8} & \textbf{50.6} & {78.0} & {86.7}\\
        
         \hline
    \end{tabular}
    \label{tab:retrieval_coco}
    \vspace{-3mm}
\end{table*}

\begin{table}[t]
\small
    \centering
    \caption{Evaluation of image-to-text retrieval (TR) and text-to-image retrieval (IR) on Flickr30K dataset.  "-" indicates the detail is not reported.}
    \vspace{-3mm}
    \renewcommand\tabcolsep{1pt}
    \begin{tabular}{c|c|ccc|ccc}
    	\hline
         \multirow{2}{*}{Model} & \multirow{2}{*}{Backbone} & \multicolumn{3}{c}{TR} & \multicolumn{3}{c}{IR} \\
         & & R@1 & R@5 & R@10 & R@1 & R@5 & R@10\\
         \hline
         \hline
         VSE++\cite{faghri2017vse++} & R152 & 52.9 & 80.5 & 87.2 & 39.6 & 70.1 & 79.5 \\
         SCAN\cite{lee2018stacked} & R101 & 67.4 & 90.3 & 95.8 & 48.6 & 77.7 & 85.2 \\
         \hline
       
         ViLBERT\cite{lu2019vilbert} & R101 & - & - & - & 58.2 & 84.9 & 91.5 \\
         Unicoder-VL\cite{li2019unicoder} & - & {86.2} & 96.3 & 99.0 & {71.5} & 90.9 & 94.9 \\
       
         UNITER\cite{chen2020uniter} & R101 & 85.9 & 97.1 & 98.8 & \textbf{72.5} & 92.4 & \textbf{96.1} \\
        
         \hline
       
         SOHO (ours) &  R101 & \textbf{86.5} & \textbf{98.1} & \textbf{99.3} & \textbf{72.5} & \textbf{92.7} & \textbf{96.1}  \\
      
         \hline
    \end{tabular}
    \label{tab:retrieval_flickr}
    \vspace{-4mm}
\end{table}
\vspace{-2mm}
\section{Experiment}
\subsection{Implementation Details}\label{subsection:implementation_details}
\vspace{-1mm}
For the \textbf{language processing}, we follow BERT to use the WordPiece tokenizer \cite{wu2016google} to split each text into language tokens. 
For the \textbf{visual processing}, as most previous works adopt feature extractor which uses $600\times 1000$ as input resolution, we also adopt setting to resize the shorter edge of input images to $600$, and limit the longer edge to be lower than $1000$ for a fair comparison.
We use pre-trained models based on public accessible ImageNet \cite{deng2009imagenet} and BERT~\cite{devlin2018bert} to initialize the parameters of our \textbf{visual backbone} and \textbf{Transformer architecture}, respectively.
Specifically, we adopt the widely-used ResNet-101 backbone and 12-layer Transformer to fairly compare with other works, while we adopt a lightweight ResNet-18 backbone and 3-layer Transformer in our ablation studies to reduce experiment cost. We will use $RX$ to denote X-layer ResNet architecture in the rest of this paper for simplicity (e.g. R101 denotes ResNet-101).
Since the visual backbone and Transformer favor different kinds of optimizers~\cite{zhang2019adam}, we follow the suggestion of Zhang \textit{et al.}~\cite{zhang2019adam} to use SGD and AdamW optimizers for visual backbone and Transformer respectively.
We use SGD with learning rate $1\mathrm{e}{-2}$ and weight decay $5\mathrm{e}{-4}$ for the visual backbone, and apply AdamW with learning rate $1\mathrm{e}{-4}$ and weight decay $1\mathrm{e}{-2}$ for Transformer.
We pre-train our model with 32 NVIDIA Tesla V100 GPUs with a batch size of $4,096$ image-text pairs.
The training process takes 40 epochs until convergence and we empirically decay the learning rate by 10 times at $25^{th}$ and $35^{th}$ epoch.

We adopt mixed-precision training to reduce memory cost and speed up training procedure. An image will be paired with four texts in each batch during pre-training, including two positive pairs and two negative pairs. We only apply MLM and MVM on the positive image-text pairs.
\vspace{-1mm}
\subsection{Downstream Tasks and Results}\label{subsection:dt_base}
\vspace{-1mm}
We test the performance of SOHO on four well-established downstream tasks, include image-text retrieval, visual question answering (\textbf{VQA}), 
natural language for visual reasoning(\textbf{NLVR}), and fine-grained visual reasoning (Visual Entailment, or \textbf{VE}).Image-text retrieval task includes two sub-tasks, i.e., image-to-text retrieval (\textbf{TR}) and text-to-image retrieval (\textbf{IR}), and are conducted on Flickr30K \cite{xie2017aggregated} and MSCOCO \cite{lin2014microsoft} datasets.
The tasks of VQA, NLVR, and VE are conducted on datasets of VQA 2.0 \cite{goyal2017making}, NLVR$^2$ \cite{suhr2019corpus} and SNLI-VE~\cite{xie2018visual} respectively.
Table \ref{table:tasks} summarizes the statistics of all our downstream tasks.

We compare our approach with several task-specific methods and pre-training models.
Most pre-training models adopt Transformer-like architectures~\cite{vaswani2017attention} with BERT-like objectives~\cite{devlin2018bert} to learn cross-modal representations~\cite{chen2020uniter,li2019unicoder,li2019visualbert,lu2019vilbert,su2019vl,tan2019lxmert,zhou2019unified}.
For downstream tasks, we find that using input features of the VD module for visual representation is better than directly applying VD embedding. We adopt the former setting in our experiment. This shows that VD suits visual representation learned with a very large scale of semantics while dense features provide more details in a relatively small dataset. 

\vspace{-2mm}
\subsubsection{Task I: Image-Text Retrieval}\label{subsection:task_itr}
\vspace{-2mm}
Image-text retrieval requires a model to retrieve the most relevant caption from candidate images, or vice versa.
It is one of the most typical tasks in the field of vision-language learning which enables a broad range of applications (e.g., image searching).
Image-text retrieval includes two sub-tasks of image-to-text retrieval (\textbf{TR})
and text-to-image retrieval (\textbf{IR}).
During training, we construct \textit{aligned} and \textit{unaligned} pairs inside of a mini-batch like most image-text retrieval models. We randomly sample $t$ aligned image-caption pairs from ground truth annotations to form a mini-batch. All the other $t-1$ captions are used as the unaligned captions for each image.
To encourage the model to predict the right labels for both the aligned and unaligned pairs, we consider the retrieval task as a binary classification problem.

In our implementation, we use the joint embedding representation of the [\textit{CLS}] token from Transformers to predict whether an image-caption pair is aligned or not.
Since the objective of image-text retrieval task is consistent with the image-text matching (ITM) task in pre-training stage, the pre-trained parameters can well be inherited for fine-tuning.
We adopt AdamW optimizer with $1\mathrm{e}{-4}$ learning rate and $1\mathrm{e}{-2}$ weight decay.
The mini-batch size $t$ is set to $24$.
We train 20 epochs until convergence and decay the learning rate by half at $3^{rd}$, $5^{th}$ , $9^{th}$ and $13^{th}$ epoch empirically.

We conduct experiments on MSCOCO~\cite{lin2014microsoft} and Flickr30k~\cite{plummer2015flickr30k}, and the results are shown in Table~\ref{tab:retrieval_coco} and Table~\ref{tab:retrieval_flickr} respectively.
It worth noting that UNITER additionally uses out-of-domain datasets and the results are expected to be better than merely use in-domain datasets as they reported~\cite{chen2020uniter}. Unicoder-VL~\cite{li2019unicoder} adopts merely out-of-domain datasets, which is also not directly comparable to our SOHO.
Nevertheless, SOHO outperforms the most recent VLPT works under most metrics on both MSCOCO and Flickr30k.
The performance improvements indicate that SOHO learns better image-text embeddings by our end-to-end pre-training architecture,
and exploits comprehensive yet compact visual semantic abstraction by the proposed visual dictionary.

\begin{table}[t]
\small
\centering
\caption{Evaluation of VQA on VQA 2.0 dataset. "-" indicates the detail is not reported. X101 denotes ResNeXt-101 architecture~\cite{xie2017aggregated}.}
\vspace{-3mm}
\begin{tabular}{c|c|c|c}
    \hline 
    Model & Backbone & test-dev & test-std \\
    \hline \hline
    MUTAN\cite{ben2017mutan} & R152 &60.17 & - \\
    BUTD\cite{Anderson_2018_CVPR} & R101 &65.32 & 65.67 \\
    \hline
    Unified VLP \cite{zhou2019unified} & X101 & 70.50 & 70.70 \\
    ViLBERT\cite{lu2019vilbert} & R101 &70.55 & 70.92 \\
    VisualBERT\cite{li2019visualbert} & R152 &70.80 & 71.00 \\
    VLBERT\cite{su2019vl} & R101 &71.79 & 72.22 \\
    LXMERT\cite{tan2019lxmert} & R101 &72.42 & 72.54 \\
    UNITER\cite{chen2020uniter} & R101 &72.70 & 72.91 \\
    \hline
    SOHO (Ours) & R101 & \textbf{73.25} & \textbf{73.47} \\
    \hline
\end{tabular}{
 \label{tab:vqa}
}
\vspace{-4mm}
\end{table}
\vspace{-4mm}
\subsubsection{Task II: Visual Question Answering}\label{subsection:task_vqa}
\vspace{-1mm}
Visual Question Answering (VQA) requires a model to take an image and a question as input and output an answer.
This task requires machines to act like humans and reason across vision and language, which is approaching intelligent AI. We model VQA as a classification problem by learning multi-layer perception from the [\textit{CLS}] token. We follow \cite{kim2018bilinear} to treat is as a $3,192$-way classification problem, and optimize the model via binary cross-entropy loss. We fine-tune for 18 epochs with a batch size of 256 until convergence.
We set the optimizer the same as in the pre-training stage.
The initial learning rates are also set the same as pre-training, and we decay the learning rate by 10 at the $12^{th}$ and $16^{th}$ epochs empirically.

Results are presented in Table~\ref{tab:vqa}.
The most direct comparable baseline to our SOHO is LXMERT~\cite{tan2019lxmert}, which adopts the same backbone and pre-training dataset as our SOHO.
SOHO obtains 0.83\% and 0.93\% absolute improvements on test-dev and test-std split over LXMERT respectively.
It is worth noting that SOHO outperforms UNITER~\cite{chen2020uniter} even under an inferior experimental setting, where UNITER additionally uses out-domain datasets in the pre-training stage.
The promising results of SOHO on VQA demonstrate that our end-to-end pre-training approach enables intelligent question answering on visual contents.

\begin{table}[t]
\centering
    \caption{Evaluation of Visual Reasoning on NLVR$^2$ dataset.}
    \vspace{-3mm}
 \begin{tabular}{c|c|c|c}
 \hline
       Model & Backbone & dev & test-P\\
       \hline \hline
       Image Only\cite{suhr2019corpus} & R152 & 51.60 & 51.90 \\
       CNN+RNN\cite{suhr2019corpus} & R152 & 53.50 & 52.40 \\
       MaxEnt\cite{suhr2019corpus} & R152 & 54.10 & 54.80 \\
       \hline
       VisualBERT\cite{li2019visualbert} & R152 & 67.40 & 67.00 \\
       LXMERT\cite{tan2019lxmert} & R101 & 74.90 & 74.50 \\
       UNITER\cite{chen2020uniter} & R101 & 75.85 & 75.80 \\
       \hline
       SOHO (Ours) & R101 & \textbf{76.37} & \textbf{77.32} \\
       \hline
    \end{tabular}
    {
    \label{tab:nlvr}
}
\vspace{-4mm}
\end{table}

\vspace{-4mm}
\subsubsection{Task III: Visual Reasoning}\label{subsection:task_nlvr}
Visual Reasoning with Natural Language (NLVR) requires a model to predict whether a text is related to a given pair of images.
Compared with VQA, NLVR addresses the challenge of compositional visual reasoning on relations, comparisons, and quantities.
We conduct this task on NLVR$^2$ dataset \cite{suhr2019corpus}.
In our implementation, we follow LXMERT~\cite{tan2019lxmert} and UNITER~\cite{chen2020uniter} to input two pairs of image-text to Transformer and get two embedding vectors from [\textit{CLS}] tokens. 
Then we learn a classifier on the concatenation of the embedding vectors over ``true'' or ``false'' by a cross-entropy loss.
The settings of the optimizer, epoch number, and learning rate are the same as VQA settings.
Since the number of input images for NLVR$^2$ is twice as VQA, the batch size of NLVR$^2$ is half of VQA.

We mainly compare with the SOTA results provided by LXMERT~\cite{tan2019lxmert} and UNITER~\cite{chen2020uniter} under the same settings for fair comparisons.
From the results shown in Table~\ref{tab:nlvr}, we observe 0.52\% and \textbf{1.52\%} absolute gains of SOHO against UNITER on \textit{dev} and \textit{test-P} split respectively.
This result validates that SOHO also has advantages when applying to compositional visual reasoning tasks.

\begin{table}[t]
\small
\centering
\caption{Evaluation of Visual Entailment on SNLI-VE.}
\vspace{-3mm}
\begin{tabular}{c|c|c|c}
    \hline 
    Model & Backbone & val & test \\
    \hline \hline
    EVE-Image\cite{xie2018visual} & R101 & 71.56 & 71.16 \\
    \hline
    UNITER\cite{tan2019lxmert} & R101 & 78.59 & 78.28 \\
    \hline
    SOHO (Ours) & R101 & \textbf{85.00} & \textbf{84.95} \\
    \hline
\end{tabular}{
 \label{tab:ve}
}
\vspace{-4mm}
\end{table}

\begin{table*}[t]
\small
    \centering
    \caption{Ablation study on the effectiveness of Visual Dictionary (VD) and the embedding vector size of VD. Results are obtained under the settings of a ResNet-18 backbone and a 3-layer Transformer architecture.
    Image-text Retrieval is conducted on the MSCOCO 1k test set.
    The top-1 and top-2 results of each metric are highlighted in bold and underlined respectively.
    Notation $\Delta$ indicates the performance gains of 2048 VD size results over baseline results without VD.
    }
    \vspace{-2mm}
    \begin{tabular}{cc|ccc|ccc|cc|cc|cc}
    	\hline
         & \multirow{2}{*}{VD size} & \multicolumn{3}{c|}{Text Retrieval} & \multicolumn{3}{c|}{Image Retrieval} & \multicolumn{2}{c|}{VQA} & \multicolumn{2}{c|}{NLVR$^2$} & \multicolumn{2}{c}{SNLI-VE} \\
         & & R@1 & R@5 & R@10 & R@1 & R@5 & R@10 & test-dev & test-std & dev & test-P & val & test\\
         \hline
         w/o VD 
         & - & 72.80 & \underline{93.20} & 96.90   & 58.22 & 88.32 & 94.40   & 66.08 & 66.33 & 62.62 & 62.61 & 82.28 & 82.16\\
         \hline
         \multirow{4}{*}{w/ VD}
         & 1024 & \underline{73.40} & 92.10 & 97.00   & \underline{58.55} & 88.84 & 94.70   & \underline{66.75} & 66.95 & 63.32 & 64.60 & \underline{82.47} & \underline{82.55}\\
         & 2048 & \textbf{75.50} & \textbf{93.50} & \textbf{97.30}   & \textbf{59.03} & \underline{88.88} & \underline{94.84}  & 66.69 & \underline{67.09} & \textbf{64.62} & \textbf{65.32} & \textbf{82.56} & 82.54\\
         & 4096 & 71.20 & \underline{93.20} & \textbf{97.30}   & 58.50 & \textbf{88.92} & \textbf{94.96}   & \textbf{66.76} & 66.91 & \underline{63.60} & \underline{64.80} & {82.53} & \underline{82.55}\\
         & 8192 & 72.10 & 92.30 & 96.50   & 58.01 & 88.08 & 94.70 & 66.65 & \textbf{67.10} & 63.15 & 64.49 & 82.29 & \textbf{82.69}\\
         \hline
         $\Delta$ & 2048 & 2.70$\uparrow$ & 0.30$\uparrow$ & 0.40$\uparrow$   & 0.81$\uparrow$ & 0.56$\uparrow$ & 0.44$\uparrow$   & 0.61$\uparrow$ & 0.76$\uparrow$ & 2.0$\uparrow$ & 2.71$\uparrow$ & 0.28$\uparrow$ & 0.38$\uparrow$ \\
         \hline
    \end{tabular}
    \label{tab:ablation}
    \vspace{-4mm}
\end{table*}

\vspace{-4mm}
\subsubsection{Task IV: Visual Entailment}\label{subsection:task_ve}
\vspace{-1mm}

Visual Entailment (VE) is a fine-grained visual reasoning task to predict whether an image semantically entails a text.
In pursuit of visual intelligence, the relationship between an image and a text pair in the VE task is more fine-grained than VQA and NLVR, which can be true (entailment), false (contradiction) or neutral.
SNLI-VE dataset~\cite{xie2018visual} is proposed for the VE task and is constructed based on Stanford Natural Language Inference (SNLI)~\cite{bowman2015large} and Flickr30K~\cite{plummer2015flickr30k} datasets.
We follow UNITER~\cite{chen2020uniter} to treat the VE task as a three-way classification problem and predict the scores of each class by a fully-connected layer on the representation of the [\textit{CLS}] token from the output of the Transformer.
We fine-tune the model for 6 epochs with batch size 128 until convergence.
The learning rate is initialized as 1e-4, and decay to 1e-5 after four epoch empirically.

We compare SOHO with a VLPT work UNITER~\cite{chen2020uniter} and a task-specific method EVE-Image~\cite{xie2018visual}.
As reported in Table~\ref{tab:ve}, SOHO achieves \textbf{85.00\%} and \textbf{84.95\%} accuracy on val and test split respectively.
The results significantly outperform the SOTA results provided by UNITER~\cite{chen2020uniter}, where \textbf{6.41\%} and \textbf{6.67\%} absolute accuracy improvements are obtained on the val and test split respectively.
The results indicate the advantage of our end-to-end framework for refining the CNN backbone together with the cross-modal Transformer to facilitate thorough vision-language alignment.

\subsection{Ablation Study}\label{subsection:ablation_study}
We perform ablation studies to validate the effectiveness of the visual dictionary (VD) on all downstream tasks.
We first establish a baseline model without VD, then incorporate VD with the baseline and further evaluate the influence of the embedding vector size (VD size) $k$.

Results are presented in Table~\ref{tab:ablation}.
Generally, we observe that for most tasks, a VD size of 2048 or 4096 achieves the best results among four sizes ranging from 1024 to 8192.
This is reasonable as VD is designed to aggregate similar visual semantics into the same image feature.
With such design, the bigger VD could learn to group more fine-grained and complete visual semantics, which benefits the VL alignment as expected.
However, too fine-grained visual semantics being grouped into different image features may deteriorate the abstraction of visual semantics, which consequently is harmful to VL alignment.
We empirically find that $k=2048$ works the best in most cases, thus we adopt $k=2048$ as our default setting.

{When compared with the baseline without VD}, our proposed method with VD enjoys better performances under almost all metrics with a wide range of $k$ (i.e., 1024, 2048, and 4096).
It validates the effectiveness of VD and shows that VD is generally applicable to a broad range of tasks.

\vspace{-1mm}
\subsection{Visualization of Visual Dictionary}
\begin{figure}
  \includegraphics[width=1.0\linewidth]{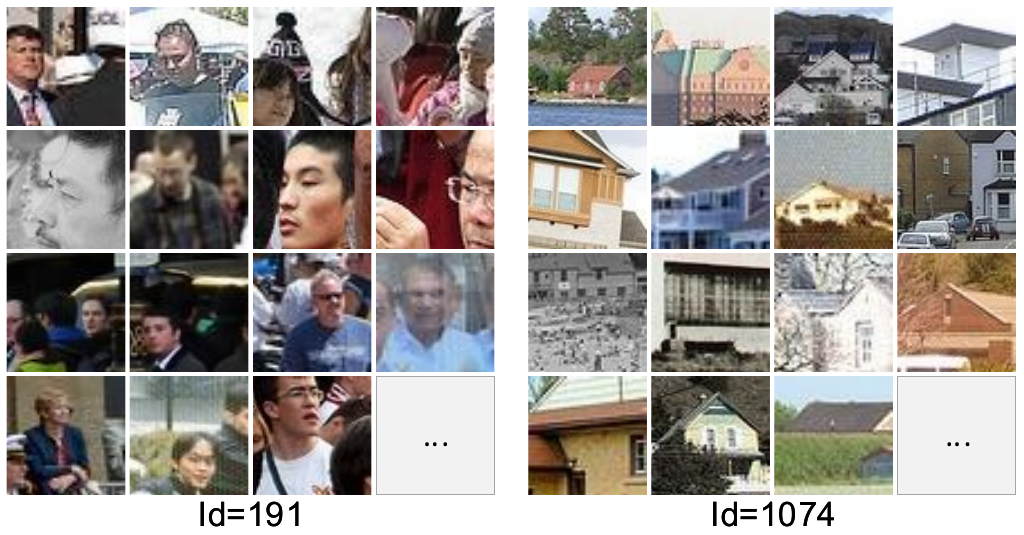}
  \vspace{-6mm}
  \caption{Visualization of VD. The left and right indices reflect the semantic of ``head'' and ``building'' with consistent visual patterns, respectively.}
  \label{fig:vis}
  \vspace{-6mm}
\end{figure}

\vspace{-1mm}
To share insights on what the proposed Visual Dictionary (VD) learned, we visualize some representative VD indices in Figure~\ref{fig:vis}.
As introduce in Sec~\ref{subsection:visual_dictionary}, a VD index is correlated with many visual features, where each visual feature corresponds to an image patch.
We randomly sample some indices from VD and visualize their corresponding image patches.
As shown in Figure~\ref{fig:vis}, the VD groups meaningful and consistent image patches into different indices, which reflects an abstraction of visual semantics. The visualization shows the strong capability of the learned VD.
More cases can be found in supplementary materials.

\vspace{-2mm}
\subsection{Inference Time}
\vspace{-1mm}
BUTD-based methods mainly include three inference stages: CNN forwarding, region feature generation, and Transformer forwarding~\cite{Anderson_2018_CVPR}.
In contrast, SOHO only includes two inference stages of CNN and Transformer forwarding.
To compare the inference efficiency of SOHO and BUTD-based methods, we set up an experiment on a V100 GPU with $600 \times 1000$ input resolution, a ResNet-101 backbone, a 12-layer Transformer, 100 boxes, 16 sentence padding length.
The average inference time for extracting BUTD features on ResNet-101 is $21$ms.
The input sequence length of the Transformer for BUTD-based methods and SOHO are $100+16=116$ and $\lceil 600/64\rceil * \lceil 1000/64 \rceil + 16=176$, respectively.
Thus the inference time of Transformer is $17$ms and $23$ms for BUTD-based methods and SOHO, respectively.
For BUTD-based methods, in addition to a $420$ms time cost of region feature generation , the main time cost, however, comes from the non-maximum suppression which s required to be applied to all $1,600$ categories.
Consequently, the $44$ms time cost of SOHO for an inference step is about $10$ times faster than the $464$ms time cost of BUTD-based methods.
Therefore, our highly-efficient SOHO could be better applied to real applications.

\vspace{-2mm}
\section{Conclusion}
\vspace{-2mm}
In this paper, we show a new perspective for vision-language model design. Particularly, we propose SOHO, one of the first end-to-end vision-language pre-training models that learns comprehensive yet compact visual representation for cross-modal understanding. To generate visual features that can be fused with language tokens, we propose a novel visual dictionary to transform an image to concrete semantics. Three pre-training tasks are conducted to build connections between images and languages. Performances on four downstream tasks show the superiority of SOHO over pre-training models with region-based image features. Moreover, we relieve the requirement for bounding box annotations, and reduce heavy human labeling costs. This end-to-end framework also shows the merit of accelerating the inference time in vision-language tasks about 10 times, which enables more online vision-language applications.
In the future, we will further explore vision-language generation tasks, and study the utilization of large-scale unpaired multi-modal data for cognition-level visual understanding.

\vspace{-2mm}
\section{Acknowledgments}
\vspace{-2mm}
This work is supported by Ministry of Science and Technology International Exchange Project (No. 43-17).

\newpage\appendix
\section{Appendix}

\subsection{Dataset Statistics}
Here we first summarize the detailed train/test image and text numbers of our pre-training and downstream datasets in Table~\ref{table:datasets}.
Then we provide a detailed comparisons of pre-training dataset usage of recent VLPT works in Table~\ref{table:pretrain_datasets}.

We follow UNITER~\cite{chen2020uniter} to classify pre-training datasets into two classes of ``in-domain'' and ``out-of-domain''.
MSCOCO Captions~ (MSCOCO)\cite{lin2014microsoft} and \textbf{V}isual \textbf{G}enome Dense Captions (VG)~\cite{krishna2017visual} are typical in-domain datasets for many VL downstream tasks (e.g., image-text retrieval).
In contrast, Conceptual Captions~\cite{sharma2018conceptual} and SBU Captions~\cite{ordonez2011im2text} are out-of-domain datasets which are noisier than in-domain datasets.
We show the dataset usage of recent VLPT works in Table~\ref{table:pretrain_datasets}.
For example, VisualBERT~\cite{li2019visualbert}, LXMERT~\cite{tan2019lxmert} and UNITER~\cite{chen2020uniter} pre-train with in-domain datasets.
Among them, UNITER~\cite{chen2020uniter} additionally use out-of-domain data for model training.
The ablation study of UNITER~\cite{chen2020uniter} shows that the additional usage of out-of-domain further improves performance.

In our work, we focus on in-domain datasets as they are commonly used in many VL tasks (e.g., image-text retrieval) and adopted by many VLPT works (e.g., VisualBERT~\cite{li2019visualbert}, LXMERT~\cite{tan2019lxmert} and UNITER~\cite{chen2020uniter}).
When comparing with UNITER, we fairly compare with its in-domain pre-training results if they are provided.
Otherwise, our ``in-domain'' dataset setting is inferior to the ``in-domain+out-of-domain'' pre-training setting of UNITER, and our results are not directly comparable.

We plan to include out-of-domain data in our pre-training data as a future work.

\begin{table}[t]
\small
\centering
\setcounter{table}{8}
\caption{Statistics of different datasets. Notation ``*'' denotes Karpathy split~\cite{karpathy2015deep}.}
\begin{tabular}{c|c|c|c|c}
\hline
\textbf{Dataset} & \multicolumn{2}{c|}{\textbf{Split}} & \textbf{\#Image (K)} & \textbf{\#Text (K)} \\
\hline

VG & \multicolumn{2}{c|}{train} & 105.9 & 472.7 \\
\hline

\multirow{4}{*}{COCO} & \multicolumn{2}{c|}{train} & 82.8 & 414.1 \\
\cline{2-5}
& \multirow{3}{*}{val} & restval* & 30.5 & 152.6 \\
\cline{3-5}
& & val* & 5.0 & 25.0 \\
\cline{3-5}
& & test* & 5.0 & 25.0 \\
\hline

\multirow{5}{*}{VQA2.0} & \multicolumn{2}{c|}{train} & 82.8 & 443.8 \\
\cline{2-5}
& \multicolumn{2}{c|}{val} & 40.5 & 214.4 \\
\cline{2-5}
& \multicolumn{2}{c|}{test-dev} & \multirow{3}{*}{81.4} & \multirow{3}{*}{447.8}\\
\cline{2-3}
& \multicolumn{2}{c|}{test-std} & & \\
\cline{2-3}
& \multicolumn{2}{c|}{test-challenge} & & \\
\hline

\multirow{3}{*}{NLVR$^2$} & \multicolumn{2}{c|}{train} & 103.2 & 86.4 \\
\cline{2-5}
& \multicolumn{2}{c|}{dev} & 8.2 & 7.0 \\
\cline{2-5}
& \multicolumn{2}{c|}{test-P} & 8.1 & 7.0 \\
\hline

\multirow{3}{*}{Flickr30K} & \multicolumn{2}{c|}{train*} & 29.0 & 145.0 \\
\cline{2-5}
& \multicolumn{2}{c|}{val*} & 1.0 & 5.0 \\
\cline{2-5}
& \multicolumn{2}{c|}{test*} & 1.0 & 5.0 \\
\hline

\multirow{3}{*}{SNLI-VE} & \multicolumn{2}{c|}{train} & 29.8 & 529.5 \\
\cline{2-5}
& \multicolumn{2}{c|}{val} & 1.0 & 17.9 \\
\cline{2-5}
& \multicolumn{2}{c|}{test} & 1.0 & 17.9 \\
\hline
\end{tabular}
\label{table:datasets}
\end{table}

\subsection{Implementation Details}
We adopt two strategies to \textbf{speed up the training procedure}.
First, we adopt mixed-precision training to reduce memory cost and speed up training procedure.
Second, we re-organize the input data in one mini-batch.
Within a mini-batch, we only forward an image once to the visual backbone if it has multiple corresponding texts, while concatenating it with each text into cross-modal transformers. For example, an image will be paired with four texts in each batch during pre-training, including two positive pairs and two negative pairs. We only apply MLM and MVM on the positive image-text pairs.

\subsection{Visualization of Visual Dictionary}
To show the semantic of visual dictionary (VD) items, we visualize the image patches that are grouped in each indices. We have shown two examples in the paper, and in the supplementary material, we randomly select ten more indices from the VD. From the visualization shown in Figure~\ref{fig:vis1}, we can find that each item in VD has meaningful and consistent semantics. In other words, our model is able to learn unified representations to represent different semantics of the image even though we do not have object bounding box annotations for supervision.

\begin{figure*}
\begin{center}
\label{fig:vis1}
\includegraphics[width=0.86\linewidth]{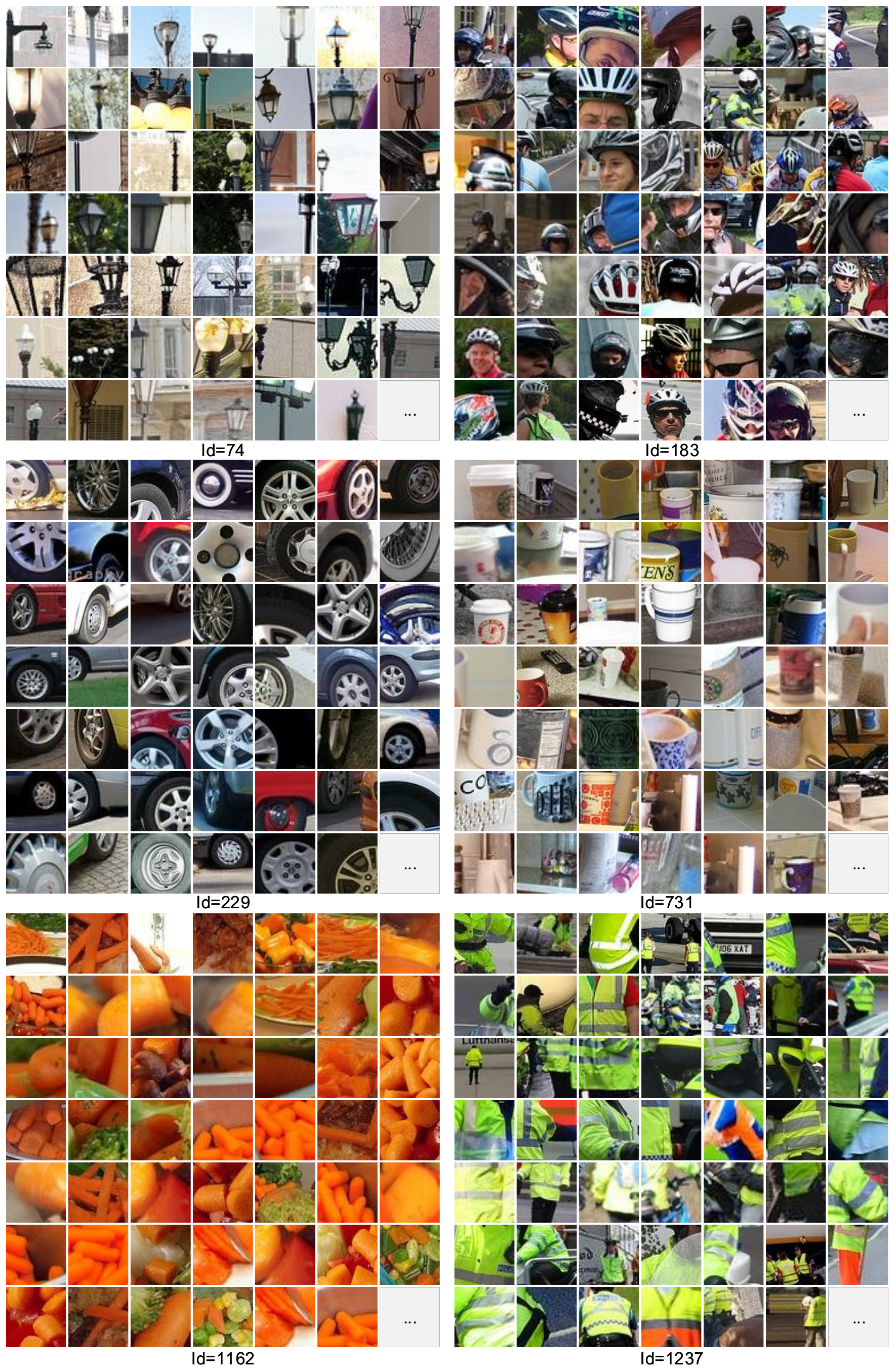}
\end{center}
\end{figure*}

\begin{figure*}
\begin{center}
\includegraphics[width=0.86\linewidth]{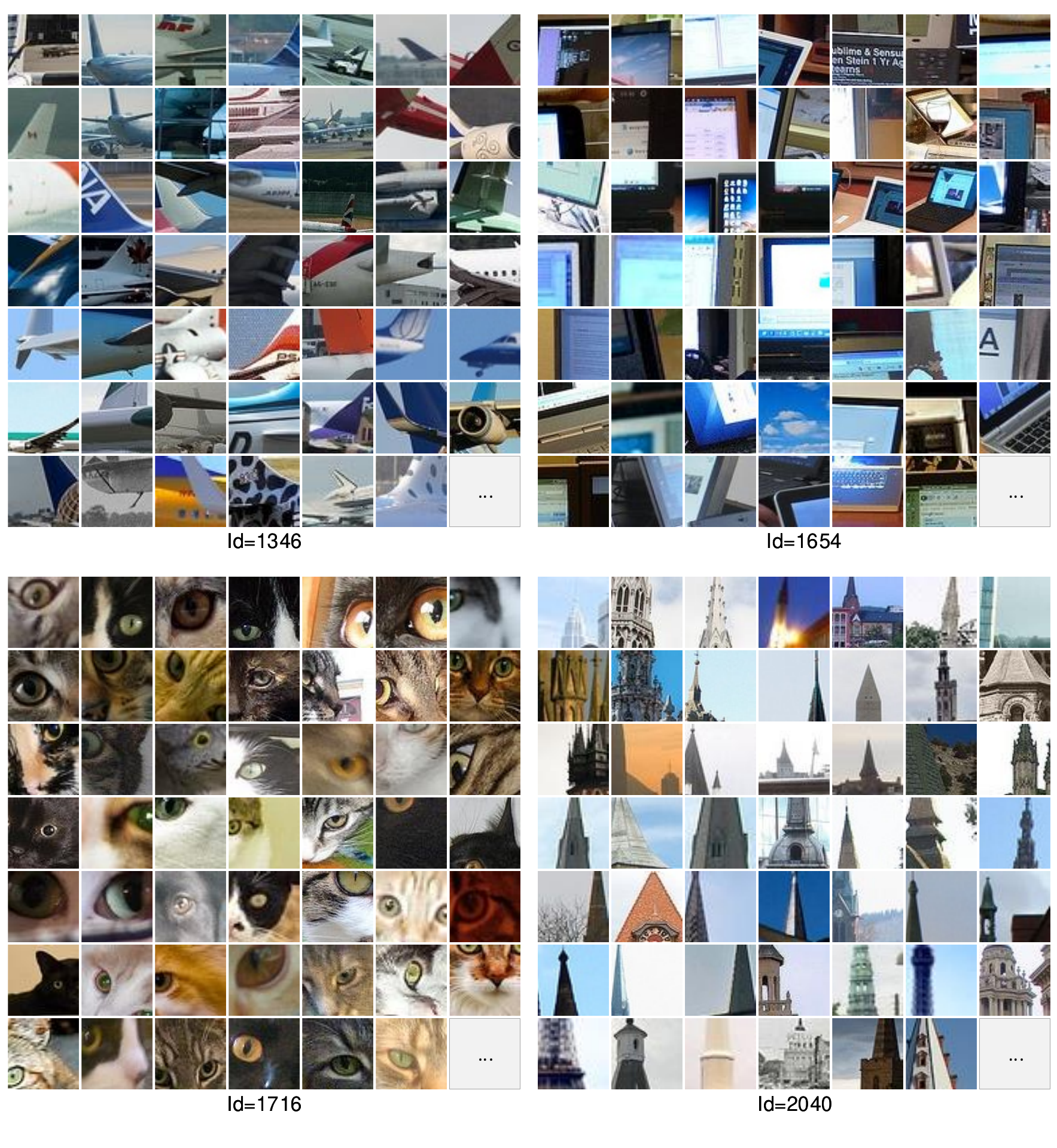}
\end{center}
\setcounter{figure}{3}
  \caption{Visualization of visual dictionary (VD) we have learned by SOHO. Apart from the two indices we have shown in the paper, we randomly select another ten indices in the visual dictionary to present in this supplementary material. From the above results we can find that, our visual dictionary is learned to group meaningful and consistent semantics of image patches into different indices. Thus, each index can reflect an abstraction of visual semantics. [Best viewed in color.]}
\label{fig:vis1}
\end{figure*}
\subsection{Discussion}

For image-text retrieval task, the traditional approaches~\cite{faghri2017vse++} first project an image and a text to a common representation space and then correlate their representations by late fusion.
For example, the widely-used late fusion method is calculating cosine similarity based on a dot-product operation, which is simple and fast.
In contrast, Transformer-based approaches early fuse the image and text by a multi-layer Transformer to get an united representation.
The unified representation captures the deep relation between an image and a text with self-attention mechanism, thus is able to achieve a better result than the late fusion representation.
However, the early fusion Transformer-based approaches cannot produce separate representation for images and texts, thus suffers from slow speed due to exhaustive computation of each possible image-text combination.
Our model as well as other vision-language pre-training models are based on Transformers, and the inference speed has become a bottleneck for applying these models to real-world search engines.
For future works, we are curious about how we could speedup the Transformer-based approaches in image-text retrieval task.
\begin{table*}[t]
\centering
\small
\caption{{Statistics on the datasets used in recent vision-and-language pre-training works.}}
\begin{tabular}{c c c c c}
\hline
& \multicolumn{2}{c}{In-domain} & \multicolumn{2}{c}{Out-of-domain} \\
\cmidrule(lr){2-3} \cmidrule(lr){4-5}
Dataset & Visual Genome~\cite{krishna2017visual} & MSCOCO~\cite{lin2014microsoft} & Conceptual Captions~\cite{sharma2018conceptual} & SBU~\cite{ordonez2011im2text}\\
Caption/Image Num & 5,060K/101K & 533K/106K & 3,000K/3,000K & 990K/990K \\
\hline
    Unified VLP~\cite{zhou2019unified} &  &  & \checkmark &  \\
    ViLBERT~\cite{lu2019vilbert} &  &  & \checkmark &  \\
    VLBERT~\cite{su2019vl} &  &  & \checkmark &  \\
    Unicoder-VL~\cite{li2019unicoder} &  &  & \checkmark & \checkmark  \\
    VisualBERT~\cite{li2019visualbert} &  & \checkmark &  &  \\
    LXMERT~\cite{tan2019lxmert} & \checkmark & \checkmark &  &  \\
    UNITER~\cite{chen2020uniter} & \checkmark & \checkmark & \checkmark & \checkmark \\
\hline
    Ours & \checkmark & \checkmark &  &  \\
\hline
\end{tabular}
\label{table:pretrain_datasets}
\end{table*}
{\small
\bibliographystyle{ieee_fullname}
\bibliography{egbib_dit}
}
\end{document}